\documentclass[5p]{elsarticle}
\usepackage{prletters}
\usepackage{algpseudocode}
\usepackage{algorithm}
\usepackage{amsmath}
\usepackage{amsfonts}
\usepackage{amssymb}

\usepackage{lineno,hyperref}
\modulolinenumbers[5]

\journal{Journal of \LaTeX\ Templates}

\usepackage{graphicx}

\begin{document}

\begin{frontmatter}

\title{Few Shots Are All You Need: A Progressive Few Shot Learning Approach for Low Resource Handwriting Recognition  }

% \address{Radarweg 29, Amsterdam}
% \fntext[myfootnote]{Since 1880.}

\author[cvcuabaddress]{Mohamed Ali Souibgui\corref{mycorrespondingauthor}}

\cortext[mycorrespondingauthor]{Corresponding author}
\ead{msouibgui@cvc.uab.cat}

\author[cvcuabaddress]{Alicia Fornés}
\author[yousriaddress,yousriaddress2]{Yousri Kessentini}
\author[uppsalaaddress]{Beáta Megyesi}


% \author[mysecondaryaddress]{Global Customer Service}

\address[cvcuabaddress]{Computer Vision Center, Computer Science Department, Universitat Aut\`{o}noma de Barcelona, Spain}
\address[yousriaddress]{Digital Research Center of Sfax, B.P. 275, Sakiet Ezzit, 3021 Sfax, Tunisia}
\address[yousriaddress2]{SM@RTS : Laboratory of Signals, systeMs, aRtificial Intelligence and neTworkS, Tunisia}
\address[uppsalaaddress]{Department of Linguistics and Philology, Uppsala University, Sweden}

\begin{abstract}
% Recognizing  scarce non-annotated data is a challenging problem in modern deep learning algorithms. The main difficulty comes  from the absence of large annotated datasets. In  a previous work, we proposed a few shot learning model that can recognize these types of manuscripts with   small amount of labeled data. But, since annotating these few data (which are in fact thousands of handwritten symbols + their bounding boxes) is still costly,  we propose in this paper a semi-supervised approach to solve the problem.  Our method is automatically assigning pseudo-labels of the non annotated   data, progressively. We show that, in different manuscript types,  our pseudo-labels can lead to  competitive results compared to using the ground truth labels with a significant  gain in human labor.

Handwritten text recognition in low resource scenarios, such as manuscripts with rare alphabets, is a challenging problem. The main difficulty comes from the very few annotated data and the limited linguistic information (e.g. dictionaries and language models). {Thus, we propose a few-shot learning-based handwriting recognition approach that significantly reduces the human labor annotation process, requiring only  few images of each alphabet symbol. The method consists in detecting all the symbols of a given alphabet in a textline image and decoding the obtained similarity scores to the final sequence of transcribed symbols}. Our model is first pretrained on synthetic line images generated from any alphabet, even though different from the target domain. A second training step is then applied to diminish the gap between the source and target data. Since this retraining  would require annotation of thousands of handwritten symbols together with their bounding boxes, we propose to avoid such human effort through an unsupervised progressive learning approach that automatically assigns pseudo-labels to the non-annotated data. The evaluation on different manuscript datasets show that our model can lead to competitive results with a significant reduction in human effort. The code will be publicly available in this repository:
\url{https://github.com/dali92002/HTRbyMatching} %labor compared to using expensive manually annotated data . }
\end{abstract}


\end{frontmatter}

% \linenumbers

\section{Introduction}
% - problem with scarce handwritten like ciphers or historical ( few pages  and unlabeled )  
% - Few shot ?
% - still need annotation
% - annotation is time consuming
% - curriculum pseudo labeling 

%Modern Handwritten Text Recognition (HTR) approaches are based on deep learning architectures. data hungry, so their performance is highly affected in low resource scenarios. 
Training data-hungry deep learning-based models in low resource scenarios is challenging due to the scarcity of labeled data. This is particularly the case of modern Handwritten Text Recognition (HTR) systems when applied to manuscripts with rare scripts or unknown alphabets. For example, ancient civilizations used no longer used alphabets (e.g. cuneiform, Egyptian hieroglyphs) and historical ciphers (used in diplomatic reports, secret societies, or private letters) often used invented alphabets to hide their contents \cite{megyesi2019decode}.

%\{}{TO DISCUSS \cite{souibgui2021one}.}

Recognizing and extracting information from these documents is important to understand our cultural heritage, since it helps to shed new light on and (re-)interpret our history  \cite{megyesi2020decryption}. However, a manual transcription is unfeasible due to the amount of manuscripts, and the automatic recognition is difficult due to the very few availability of annotated data for training. Moreover, the problem becomes harder in ciphers because when the alphabet is invented, no dictionaries or language models are available. 

Contrary to deep learning models, human beings are able to learn new concepts from one or few examples. To imitate this ability, recent research is being conducted in the field called few-shot learning \cite{wang2020generalizing}. For this reason, we explored in our previous work \cite{souibgui2020} whether few shot detection could be adapted for recognizing enciphered manuscripts. The main reason is that typical HTR models must be trained on the particular alphabet to be recognized, and whenever the alphabet changes, the system must be retrained from scratch with samples from the new script. For this reason, we treated the recognition as a symbol detection task: by providing one or few examples of each symbol alphabet, the system was locating them in the manuscript. Therefore, the model was able to cross multiple scripts, while requiring only few labeled data from each new cipher alphabet. The first experimental results obtained good performance in enciphered manuscripts compared to the typical methods, while reducing the amount of labelled data for fine-tuning.

Nevertheless, the required labelled data in our few-shot model still implied a significant human effort: labelling few enciphered pages for fine tuning means the manual transcription of thousands of symbols together with their corresponding bounding boxes. To alleviate such a problem, in this paper, we minimize such manual labeling stage by proposing an {unsupervised learning approach} % \cite{bengio2009curriculum} 
that can automatically and progressively annotate the data by assigning $pseudo$-$labels$ from the unlabeled handwritten text lines. As a result, our method requires only few-shots of the desired alphabet: the user just crops few examples (preferably 5) from each symbol to perform the pseudo-labeling, avoiding to annotate text lines (including symbol bounding boxes) from the same alphabet for fine tuning. This means that pseudo-labeled data is automatically obtained to fine-tune our model, with a zero manual effort. 

The main contributions of our work are as follows: $(i)$: We propose a few-shot learning model for transcribing manuscripts in low resource scenarios with a minimal human effort: it only requires labeling five examples from each new symbol alphabet, instead of labeling entire text lines. $(ii)$: We propose an unsupervised, segmentation-free method to progressively obtain pseudo-labelled data, which can be applied to cursive texts with touching symbols. $(iii)$ We propose a generic recognition and pseudo-labeling model that can be applied across different scripts. $(iv)$: We demonstrate the effectiveness of our approach through extensive experimentation on different datasets, reaching a performance similar to the one obtained with manually labelled data.

% The rest of this paper is organized as follows. We overview the related 
% work in Section~\ref{sec:related}. Then, we introduce our proposed method in Section~\ref{sec:method}. Afterwards, we present the performed 
% experiments and analyze the obtained results in Section~\ref{sec:experiments}. Finally, a brief conclusion is given in Section~\ref{sec:conclusion}.

% Low resource handwritten text recognition is an important yet hardly addressed problem in document analysis community. In such scenarios, the available data is not only unlabeled, but also scarce. Ciphered manuscripts is a type, perhaps the most representative, of this problem. Because, invented alphabets were replacing the original ones to hide a manuscript content. 

% The invented alphabets, called cipher alphabets, are hard  to track, because they are changing between different enciphered manuscripts. This complicates  its recognition  using the recently successful in Handwritten Text Recognition (HTR), deep learning tools. Since there is no language models or dictionaries, rather than the usual tiny amount of  documents (Only one or few pages for each specific ciphered record).  In \cite{souibgui2020}, 

\begin{figure*}[t!]
    \centering
    \includegraphics[width=1\linewidth,height=5.5cm]{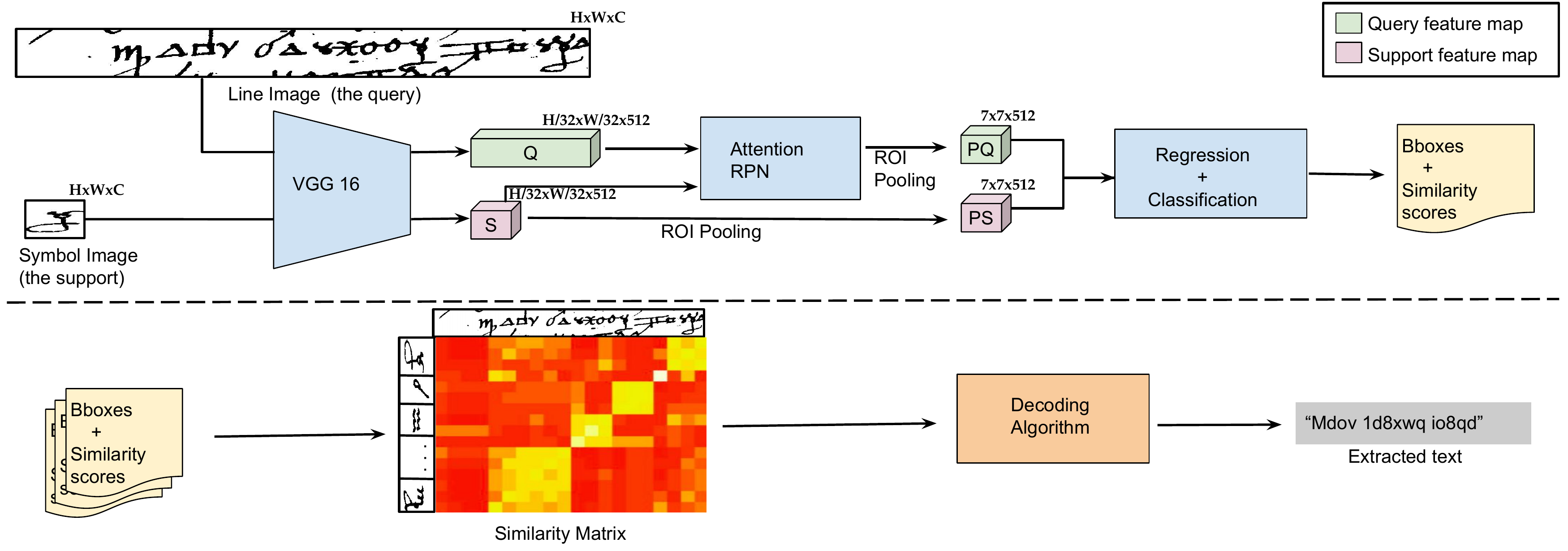}%,height=4.5cm
    \caption{{Our few-shot approach for handwriting recognition. Examples of each symbol in the alphabet are used as supports. Up: Detection of a support symbol in a handwritten line. Down: Construction of the similarity matrix from the predicted bounding boxes and decoding it to obtain the final text.}}
    \label{fig:icprarchitecture}
\end{figure*}

\begin{figure}[t!]
    \centering
    \includegraphics[width=1\columnwidth]{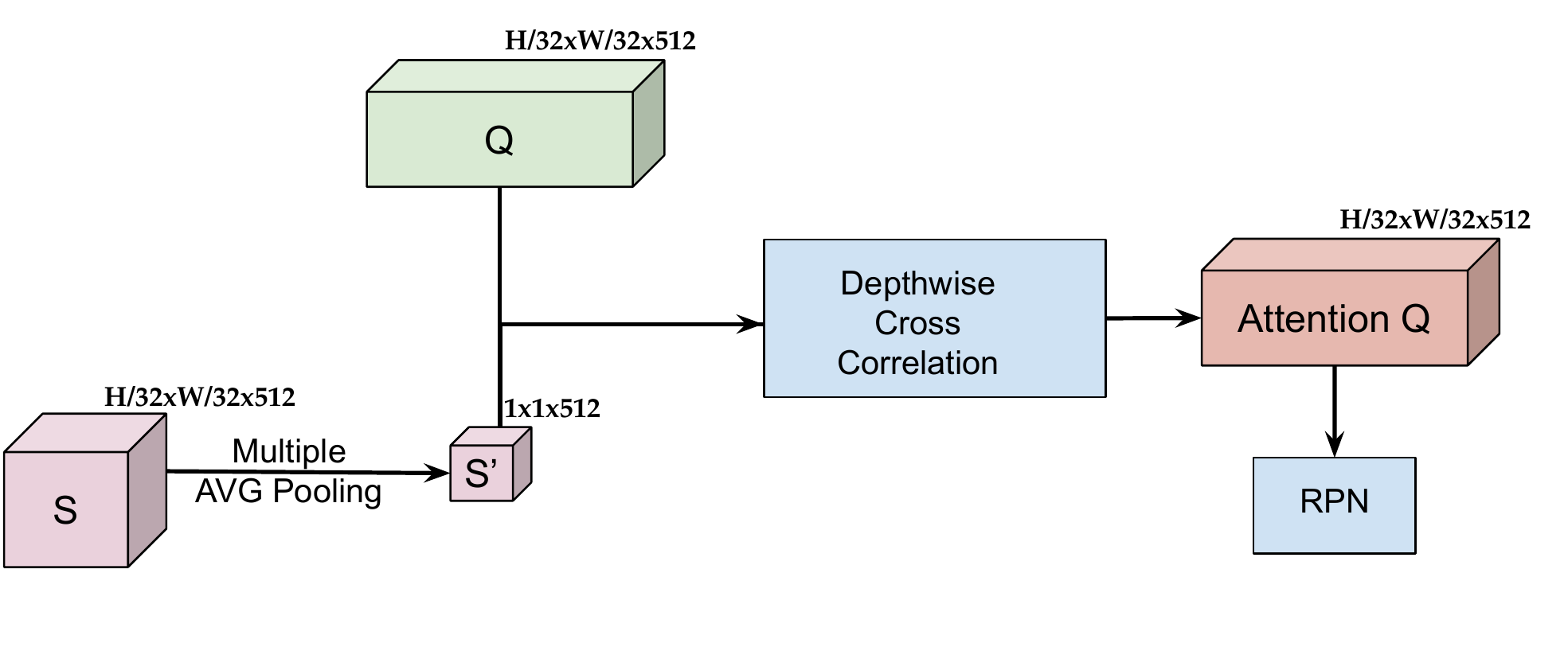}%,height=4.5cm
    \caption{{An illustration of the attention RPN: the support feature map is average pooled   until obtaining a shape of $1\times1\times512$. This later is multiplied over depth with the Query feature map to obtain the attention Q. Which is passed to the RPN for region proposing.}}
    \label{fig:attentionrpn}
\end{figure}

\section{Related Work}\label{sec:related}

\subsection{Low Resource Manuscript Recognition}

%Low resource manuscript related to historical documents is an active  field in HTR. 
Low resource handwritten text recognition applied to historical manuscripts is an active field in the document analysis community. However, the research on the transcription of enciphered manuscripts with invented alphabets is quite recent. The first attempt towards transcribing this type of handwritten text was proposed in \cite{fornes2017transcription} using MultiDimensional Long Short-Term Memory (MDLSTM) Recurrent Neural Networks \cite{graves2012offline}. The performance was satisfactory, but at the cost of a time-consuming data labeling effort. Of course, for each new cipher alphabet, a similar annotation stage was required. For this reason, some unsupervised methods were introduced \cite{baro2019,yin2019}. In those approaches, the enciphered document is first segmented into isolated symbols, then a clustering algorithm is applied to group the visually similar symbols. The main drawback of such methods is the segmentation stage, because the symbol segmentation was often inaccurate, provoking transcription errors. Similarly, and given the lack of labelled data, some researchers have opted for learning-free symbol spotting approaches \cite{rothacker2015retrieving,bogacz2016segmentation} for ancient manuscripts (e.g. Egyptian hieroglyphs, cuneiform, or runes). 

In summary, supervised methods obtain good performance but they require a large amount of labeled data. Contrary, unsupervised or learning-free methods can be applied when labelled data is not available, but they lead to a lower performance. Thus, to maintain good performance while reducing the manual annotation, few-shot learning for manuscript recognition seems preferable \cite{souibgui2020}, since it reaches a performance similar to supervised methods but requiring few annotated text lines. A similar approach based on character matching was proposed in \cite{zhang2020}, although the experiments were mostly carried out on synthetic data, instead of on real historical or cursive manuscripts.

%We showed that our model leads to a good performance using only a few annotated lines, compared to the previous HTR tools

\subsection{Handwritten Text Pseudo-Labeling}

Pseudo-labeling models aim to make profit from unlabeled data when training. In 
semi-supervised learning \cite{zhu2005semi,rasmus2015semi},  few labeled data is used to start the process. For instance, in the label propagation approach based on distances \cite{jaakkola2002partially, weston2012deep}, labels are assigned from the unlabeled data (called pseudo-labels) to be used to reinforce the training. Similarly, in \cite{lee2013pseudo}, the training started with true labels and gradually increased with pseudo labels.  
% An algorithm called MixMatch was proposed in  \cite{berthelot2019mixmatch}    basing on data augmentation, of both labeled and unlabeled data. The labels of these latter are guessed and the whole augmentations are mixed and used in the model. 
In \cite{wu2019progressive} a shared backbone extracted features from the labeled, pseudo-labeled and unlabeled data at each iteration. Then, from the feature space, the reliable labels were estimated according to the distance with the true labels while the non trusted labels were pushed away with an exclusive loss. Moreover, a pseudo-labeling curriculum approach for domain adaptation \cite{choi2019pseudo} used a density-based clustering algorithm. The idea was to annotate data with the same labels set, but taken from a different domain.

In HTR, this strategy was hardly applied mainly due to the difficulties in character segmentation, since touching characters are common in cursive texts. In \cite{frinken2012semi}, labels were guessed at word level using keyword spotting. A confidence score was used to assign new labels to the retrieved words and enlarge the dataset. Furthermore, a text to image alignment was proposed in  \cite{leifert2020two} following this strategy. 

% Curriculum learning \cite{bengio2009curriculum}, is a training strategy that feeds the model in a meaningful order (going from easiest samples, to the hardest ones). 

\section{Proposed  Approach}\label{sec:method}

In this section we describe our approach  for few-shot handwritten  text recognition. First, our model is trained on synthetic data, i.e. text line images created using various Omniglot symbol alphabets \cite{lake2015human}. Afterwards, the model is fine-tuned using the pseudo-labelling approach with the specific alphabet from the target domain (real manuscript). These steps are described next. 
% Thus we retrain it with an other set of synthetic lines constructed from the few shots (the support set) that is  the only required input of our approach to reduce human effort. Then, it can be used to predict pseudo-labels, progressively, as will be explained below.    

\subsection{Few-shot Manuscript Matching}

As stated before, few shot object detection has shown to be suitable for recognizing manuscripts in low resource scenarios. Formally, in few-shot detection, if the size of the alphabet is $N$, and  we provide $k$ examples from each symbol alphabet (named $shots$ (or supports)), the task is considered as an $N$-way $k$-shot detection problem. In such setting, the model can be trained on certain alphabets with sufficient labelled data, and later, tested on new alphabets (classes) with few labeled data. 

Our few-shot learning model, illustrated in  Fig.~\ref{fig:icprarchitecture}, is segmentation free and works at line level. As input, it takes the text line image with an associated alphabet in the form of isolated symbol images. In this step, one or few examples (usually up to five) of each alphabet symbol should be given. 
The two inputs are propagated in a shared backbone to get the feature maps. The feature maps are  used in the Region Proposal Network (RPN) with an attention mechanism, which performs the  depth-wise cross correlation between them, {as illustrated in Fig.~\ref{fig:attentionrpn}}. The Region of Interest (ROI) pooling is applied to the RPN proposals and the support image to provide well cropped symbol image candidates. Thus, we obtain two feature maps representing the regions that are candidates to contain the support image. Those are combined together and passed to the final stage where the bounding boxes are produced with the class 1 (similar to the support) or 0 (different from the support symbol). For each labeled bounding boxes, a  confidence score between 0 and 1 is predicted according to the similarity degree with the support image. We repeat this process for the all supports (all the alphabet symbols)  and  take only the  bounding boxes with high confidence score (higher than a given threshold) to construct a similarity matrix between the symbol alphabet and the line image regions. This matrix is the input of the decoding algorithm, which provides the final transcription. 

\subsection{Similarity Matrix Decoding}

\begin{figure*}[t!]
    \centering
    \includegraphics[width=\linewidth]{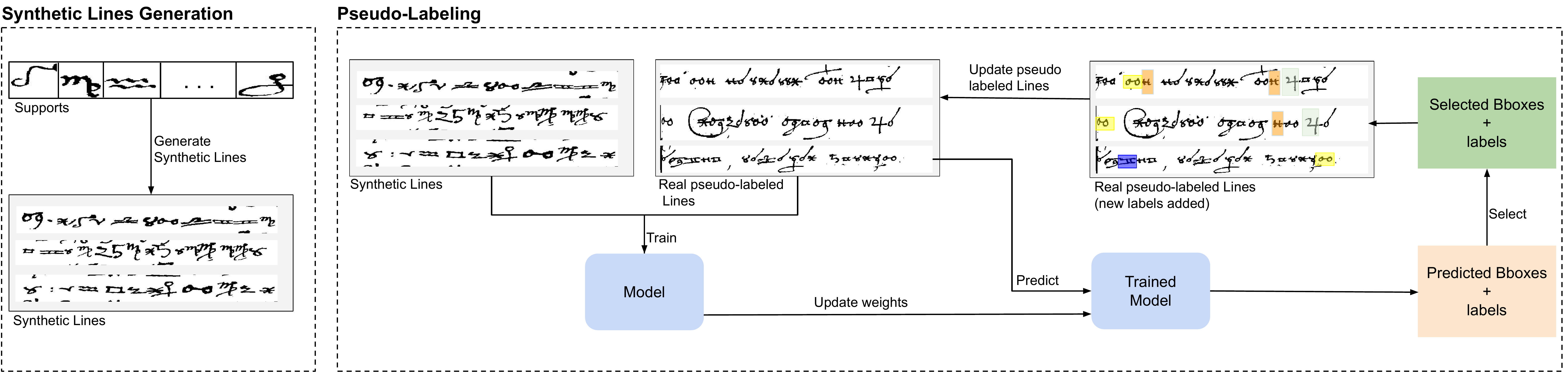}%,height=4.5cm
    \caption{Our pseudo labeling approach: At the beginning synthetic lines are generated using the supports set. Then, the pseudo-labeling phase starts. At starting, there is no pseudo-labeled data, so only synthetic lines will be used for retraining the model. Then, the model predicts symbols from the real unlabeled lines with the same script. The symbols with highest confidence score, namely pseudo-labels, are labelled and added with their predicted bounding boxes. Next, the model is retrained  again using the synthetic lines and the pseudo-labeled symbols from real lines. The process is repeated until the full dataset is annotated.}
    \label{fig:proposed_method}
\end{figure*}

The decoding algorithm shown in  Fig.~\ref{fig:icprarchitecture} takes the similarity matrix, traverses the columns from left to right,  and decides, for each pixel column, the final transcribed symbol class among the candidate symbols. Concretely, for each time step, it chooses the symbol having the maximum similarity score. To minimize errors, a symbol is only transcribed if its bounding box is not overlapped by another symbol with a higher similarity value for a certain number of successive pixels (in our case, we used 15 pixels as a threshold). Despite its simplicity, this decoding method presented in Algorithm~\ref{alg:SMD} is effective for transcribing sequences of symbols. It can be considered also as a modified version of the Connectionist Temporal Classification (CTC) algorithm \cite{graves2006connectionist}.

\begin{algorithm}[h]
\caption{{Similarity Matrix Decoding}}\label{alg:SMD}
\begin{algorithmic}
\Require 
\\ $M$ \Comment{Similarity matrix}
\\ $rep$\_$thresh$ \Comment{Repetition threshold}
%\\ $AlphabetList$ 
\Ensure $CharList$ \Comment{Characters sequence}
\Statex
$last$\_$max$ $\gets [-1,0]$ \Comment{$[index,score]$}\\
$repetition$ $\gets 0$\\
$maximums$ $\gets MaxInd(M)$ \Comment{maximum index and score for each column}\\
$CanAdd\gets False$\\
\For{\texttt{$maxi$ in $maximums$}}
        \State 
            \If{$maxi\neq last$\_$max$}
                \State $repetitions \gets 0$
                \State $CanAdd \gets True$
            \Else
                \If{$repetitions > rep$\_$thresh$ \texttt{and} $CanAdd$}
                    \State $CharList \gets CharList \cup maxi[index]$
                    \State $CanAdd \gets False$
                \Else
                    \State $repetitions \gets repetitions + 1 $
                \EndIf
            \EndIf 
\EndFor

\end{algorithmic}
\end{algorithm}

As mentioned before, our few-shot model is first trained on the Omniglot dataset: we synthetically construct lines to learn the matching in different alphabets. Then, at testing time, it can be used to  recognize unseen alphabets, requiring only a support set composed of few examples of each symbol alphabet. However, in our previous work \cite{souibgui2020}, experiments showed that the predictions can be significantly improved when we fine-tune the model using some real text lines, because there is a domain difference between the synthetic Omniglot symbols and the real historical symbols.

\subsection{Progressive  Pseudo Labeling}\label{sect:pseudo_labeling}

\begin{figure}[t!]
    \centering
    \includegraphics[width=\linewidth]{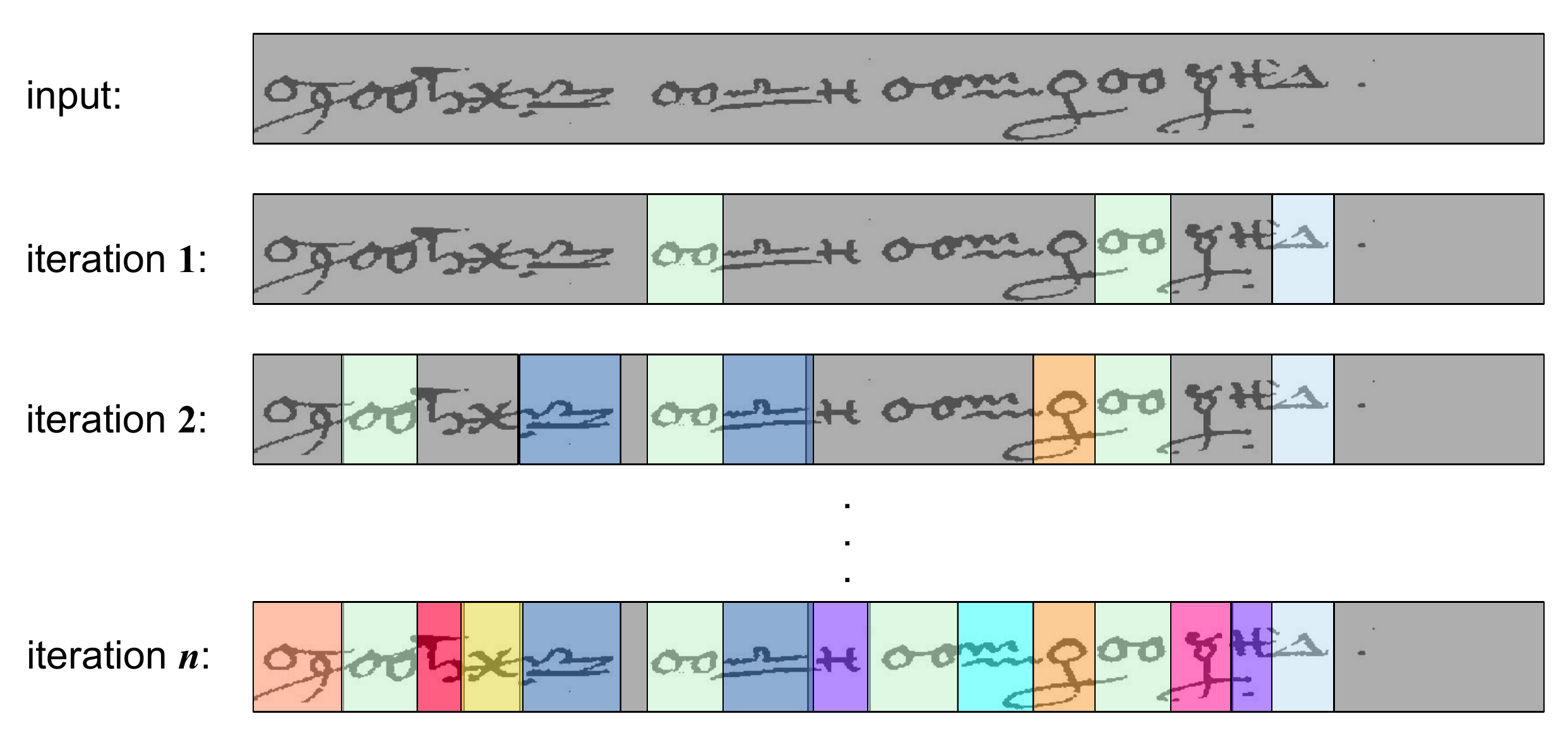}%,height=4.5cm
    \caption{An example of pseudo labeling a line image. The background is colored in grey, while the predicted label classes at each time are shown in colors. Each symbol class is shown with a different color (best viewed in color). }
    \label{fig:labeling_example}
\end{figure}

Since low resource manuscripts are mostly unlabeled, the user must provide the label of each symbol together with their corresponding bounding box. Thus, to reduce the human labor, we propose to automatically annotate the manuscripts that will be used for fine tuning the model described above. Our proposed progressive data pseudo-labeling strategy consists in the following two stages.

\subsubsection{Synthetic Data Generation}

Our few-shot model needs to be fine-tuned using data from the target domain (often with an unseen alphabet) to reduce the gap between the source and target domains. But, since we aim to minimize the user effort, we restrain the demands on a support set of few examples from each new symbol alphabet. Hence, the user must only select up to 5 samples per symbol, called shots. From those shots, we automatically generate synthetic lines by randomly concatenating them in a line image. {We tried to make those synthetic lines as realistic as possible. To do so,  the space between characters was  chosen randomly between 0 and 30  pixels, also, before concatenation  we rotate each character randomly between -5 and 5 degrees.   Moreover,  we add some artifacts to the upper part and lower part of the line to simulate a realistic segmentation of a handwritten line.} Those created lines compose our starting labeled set. Since our model was only pretrained on a different data domain, i.e. the synthetic Omniglot lines, this technique significantly improves the model prediction for unseen alphabets or scripts.

\subsubsection{Pseudo-Labeling Process}

After retraining our model with synthetic lines, we begin labeling the non-annotated data. The process is illustrated in Fig.\ref{fig:proposed_method}. Of course, at the beginning, the pseudo labeled set is empty (no labels are available), so only the synthetic lines can be used for training. Then, we pass the real text lines through our model to get the predictions, which include the bounding boxes of the regions that are similar to the input alphabet images as well as the assigned similarity score. Since the higher score, the more credible the label, we choose the top scored predictions as pseudo labels at this iteration. We experimentally found that the best option is to choose, at each iteration, the 20 \%  of the training data size as the number of the new pseudo-labels. The obtained pseudo-labeled set will be joined to the synthetic set for the next training iteration. 

This process is repeated until annotating the whole unlabeled set (all text lines), or in the case where it is not possible to add new pseudo labels with credible confidence score (we set a threshold of 0.4 as the minimum confidence score for assigning pseudo-labels). In fact, whenever the score is below this threshold, it is better not to label the symbol. Note that we label the handwritten lines without the need of segmenting them into isolated symbols. In this way, the remaining unlabeled symbols in the different lines at each iteration are considered as background during the next training. Fig.~\ref{fig:labeling_example} shows an example of a handwritten line during the pseudo labeling process. At the beginning, the whole image is considered as a background. Then, the symbols with higher confidence score are labeled in the first iteration, while the hardest ones will be labelled in the next iterations.    

% A formal explanation can be found in the following algorithm. 

\section{Experimental Results}\label{sec:experiments}
In this section we present the experiments that we performed to validate our approach. We begin by presenting the different datasets (corresponding to low resource scenario), and then, we present and discuss the results. 

\subsection{Datasets}

As low resource handwritten text, we chose historical enciphered manuscripts and Codex  Runicus.
\subsubsection{Enciphered Manuscripts}

As we said before, ciphers are a typical form of low resource  handwritten data. Many ciphers use a large variety of invented symbols instead of using common alphabets. In this work we choose two enciphered manuscripts, namely Borg and Copiale. Both are described in detail and publicly  accessible in these links\footnote{\url{https://cl.lingfil.uu.se/~bea/copiale/}}\footnote{\url{https://cl.lingfil.uu.se/~bea/borg/}}. In our experiments, we exclude the symbols with very few occurrences (once or twice), so we use 24 symbols from the Borg manuscript and 78 from the Copiale one.  Fig.~\ref{fig:datasets} shows some examples of these handwritten ciphers. As it can be observed, in the Borg cipher, the symbol segmentation is difficult because of the frequent touching symbols, which is one of our main motivations for our segmentation free proposed method. For Copiale, the size of the alphabet is large, so it can be used to test our approach for higher number of classes. We took few pages of each document for fine tuning, i.e. to perform the progressive pseudo labeling process. Also, a performance comparison when using manually produced labels is provided.  

\subsubsection{Codex  Runicus}
\begin{figure}[t]
    \begin{center}
    \begin{tabular}{c}
    % \small
    \multicolumn{1}{l}{\includegraphics[width=0.9\columnwidth, height=1.7cm]{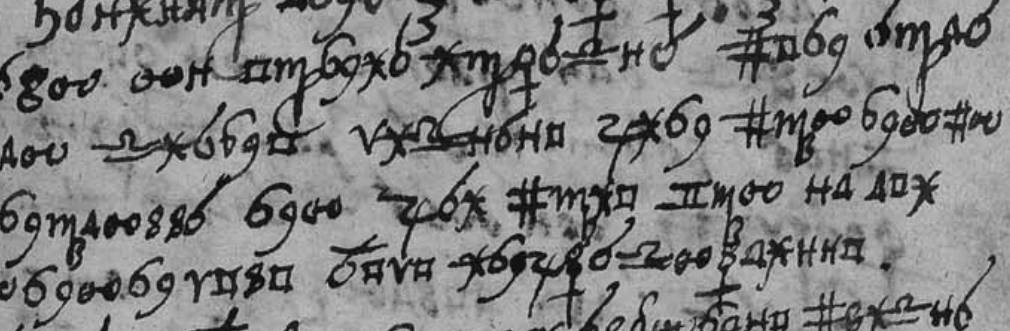}}\\
    The Borg Cipher \\
    %   \hline
    \noalign{\smallskip}
    \multicolumn{1}{l}{\includegraphics[width=0.9\columnwidth, height=1.7cm]{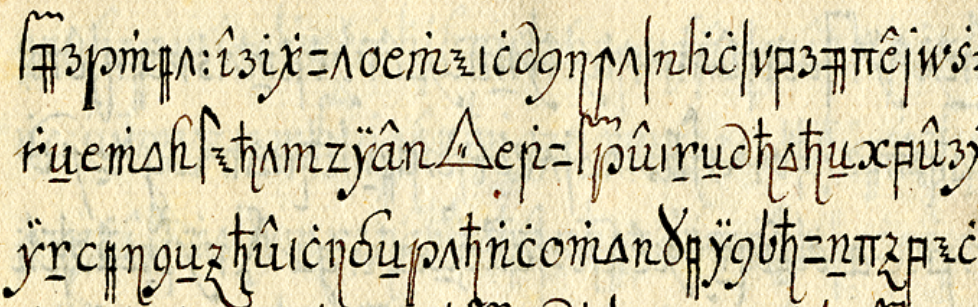}}\\
    The Copiale Cipher\\
    %   \hline
     \noalign{\smallskip}
      \multicolumn{1}{l}{\includegraphics[width=0.9\columnwidth, height=1.7cm]{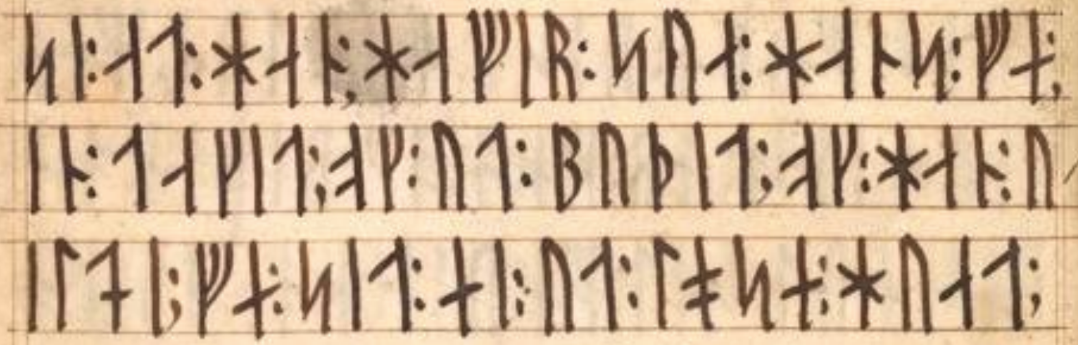}}\\
      Codex Runicus
    \end{tabular}
    \end{center}
    \caption{Examples of the three manuscripts with low resource annotated data.}
    \label{fig:datasets}
\end{figure}

The Codex Runicus is a historical  manuscript, written  on 100 parchment folios (leaves) around 1300AD in the province of Scania, in medieval Denmark. We took 10 pages to perform our experiments. Those pages were transcribed by an expert to compare it with our automatic labeling. An example of this manuscript is illustrated in Fig.~\ref{fig:datasets}. We chose this manuscript because it uses a rare alphabet and perfectly fits in our low resource handwriting recognition problem.

\subsection{Experimental Setup and Metrics}

To carry out the experiments, we first trained our proposed few-shot handwriting recognition model using lines created from the Omniglot dataset only. Then, we retrained the model using synthetic lines, created by randomly concatenating the 5 selected symbols (shots), and applying some transformations (including rotation, resizing, thickness modification, etc), hence called Synthetic Data (SD). This step is performed to reduce the domain gap between the Omniglot lines and the real lines. Afterwards, we start predicting the labels and obtaining the Pseudo Labeled Data (PSD) by using the approach detailed in Subsection~\ref{sect:pseudo_labeling}. We finally fine-tune the model with this data and compare its performance to a method that uses Real Labeled Data (RLD) for training. The evaluation is done according to the Symbol Error Rate (SER) metric. It is the same as the Character Error Rate used in HTR. Formally,  $SER = \frac{S+D+I}{N}$, where $S$ is the number of substitutions, $D$ of deletions, $I$ of insertions and $N$ is the ground-truth's length. Obviously, the lower the value, the better performance.

We compare our approach with our previous few-shot model \cite{souibgui2020}, the unsupervised \cite{baro2019,yin2019} and supervised  \cite{fornes2017transcription} approaches for ciphered manuscript recognition. 

\subsection{Results}

\begin{table*}[h]
\centering
\caption{Obtained Results on the Borg dataset. FT: Fine Tuning. Om: Omniglot. SD: Synthetic Data. RLD: Real Labeled Data. PLD: Pseudo Labeled Data. ULD: UnLabeled Data.}
\label{tab:result_borg}   
\begin{tabular}{ccccc}

\noalign{\smallskip}\hline\noalign{\smallskip}
Dataset&Method&User Effort & Training $\rightarrow$ FT & SER \\
\noalign{\smallskip}\hline\noalign{\smallskip}
\multirow{7}{*}{Borg} &Unsupervised \cite{yin2019}&None & ULD & 0.57 \\
&Unsupervised  \cite{yin2019}&Manual Segmentation &  ULD & 0.22 \\
&Unsupervised  \cite{baro2019}&Clusters Processing &  ULD & 0.54 \\
&MDLSTM \cite{fornes2017transcription}&Manual Labeling &  RLD & 0.55 \\
&Few-shot \cite{souibgui2020}&Manual Labeling & Om $\rightarrow$ RLD & 0.21 \\
&Few-shot \cite{souibgui2020}&5 shots & Om $\rightarrow$ NONE & 0.53 \\
% provide 5 shots & Omniglot & BS & 0.28 \\
% Ours&5 shots & Om $\rightarrow$ SD $\rightarrow$ PLD & 0.26 \\
&Ours&5 shots & Om $\rightarrow$ SD + PLD & 0.24 \\
\noalign{\smallskip}\hline\noalign{\smallskip}

\multirow{7}{*}{Copiale} &Unsupervised \cite{yin2019}&None & ULD & 0.44 \\
&Unsupervised  \cite{yin2019}&Manual Segmentation &  ULD & 0.37 \\
&Unsupervised  \cite{baro2019}&Clusters Processing &  ULD & 0.20 \\
&MDLSTM \cite{fornes2017transcription}&Manual Labeling &  RLD & 0.07 \\
&Few-shot \cite{souibgui2020}&Manual Labeling & Om $\rightarrow$ RLD & 0.11 \\
&Few-shot \cite{souibgui2020}&5 shots & Om $\rightarrow$ NONE & 0.39 \\
% provide 5 shots & Omniglot & BS & 0.28 \\
% Ours&5 shots & Om $\rightarrow$ SD $\rightarrow$ PLD & 0.26 \\
&Ours&5 shots & Om $\rightarrow$ SD + PLD & 0.15 \\
\noalign{\smallskip}\hline\noalign{\smallskip}

\multirow{3}{*}{Codex Runicus} %&Unsupervised \cite{yin2019}&None & ULD & 0.57 \\
% &Unsupervised  \cite{yin2019}&Manual Segmentation &  ULD & 0.22 \\
&Unsupervised  \cite{baro2019}&Clusters Processing &  ULD & 0.06 \\
&MDLSTM \cite{fornes2017transcription}&Manual Labeling &  RLD & 0.26 \\
&Few-shot \cite{souibgui2020}&Manual Labeling & Om $\rightarrow$ RLD & 0.05 \\
&Few-shot \cite{souibgui2020}&5 shots & Om $\rightarrow$ NONE & 0.40 \\
% provide 5 shots & Omniglot & BS & 0.28 \\
% Ours&5 shots & Om $\rightarrow$ SD $\rightarrow$ PLD & 0.26 \\
&Ours&5 shots & Om $\rightarrow$ SD + PLD & 0.09 \\
\noalign{\smallskip}\hline\noalign{\smallskip}

\end{tabular}
\end{table*}

In the Borg cipher, 117 non annotated training lines, containing 1913 symbols, are used to learn the pseudo-labels, whereas 273 lines are used for testing.
This manus\-cript is considered a hard case because of the overlapping symbols, which makes predicting correct bounding boxes challenging. Also, the writing style is variable. From the training lines, we crop 5 examples of each Borg symbol class. 
%(we used the same as \cite{souibgui2020}, including  24 classes/symbol types in total). 

The obtained results are shown in Table~\ref{tab:result_borg}. As it can be seen, using a few-shot method with real labels leads to a SER of 0.21, being considered as the upper bound. But, this result is costly, since a user must manually annotate 1913 symbols, including their labels and bounding boxes. We  also notice that the supervised MDLSTM with a larger training set, annotated at line level (no bounding boxes are required), obtains a moderate result, because of this manuscript's difficulties metioned before. We notice that the unsupervised methods are only useful when the segmentation of lines into isolated symbols is accurate, which is a costly task as well. Our few-shot model, trained only on Omniglot and tested on Borg, leads also to a poor result (a SER of 0.53). Of course, the reason is the difference between the training and test domains. On the other side, when using the pseudo-labeled data provided by our approach, we obtained an acceptable result of 0.24 SER, with a high gain in user effort because we only require 5 examples of each symbol, avoiding a costly manual annotation.  

The Copiale manuscript contains easy to segment symbols but with a larger alphabet size.  As it can be noticed from Table~\ref{tab:result_borg}, 
the MDLSTM performs better in this dataset because of the larger labelled training lines and a unique handwriting style. However, our model achieves a competitive result by using less data (the few-shot model is trained with 176 lines containing 7197 symbols). Anyway, annotating these lines is costly, so a better choice is to automatically  produce 
pseudo-labels. By using our pseudo-labeling process, we reach a competitive performance, compared to the manually labeled data (a SER of 0.15 versus 0.11). 

% \begin{table}[h]
% \centering
% \caption{Obtained Results on the Copiale dataset. FT: Fine Tuning. Om: Omniglot. SD: Synthetic Data. RLD: Real Labels Data. PLD: Pseudo Labeled Data.}
% \label{tab:result_copiale}   
% \noalign{\smallskip}

% \begin{tabular}{cccc}
% \hline\noalign{\smallskip}
% Method&User Effort & Training $\rightarrow$ FT & SER \\
% \noalign{\smallskip}\hline\noalign{\smallskip}
% \cite{souibgui2020}&Man. Labeling & Om $\rightarrow$ RLD & 0.11 \\
% \cite{souibgui2020}&5 shots & Om $\rightarrow$ NONE & 0.39 \\
% % provide 5 shots & Omniglot & BS & 0.28 \\
% Ours&5 shots & Om $\rightarrow$ SD $\rightarrow$ PLD & 0.15 \\
% Ours&5 shots & Om $\rightarrow$ SD + PLD & 0.15 \\
%  \noalign{\smallskip}\hline\noalign{\smallskip}
% \end{tabular}
% \end{table}

Finally, we tested our method on the  Runicus manus\-cript, as an example of ancient document with a rare alphabet. This manuscript can be considered easier than ciphers because the symbol segmentation is easy and the alphabet size is moderate. For this reason, an unsupervised clustering method can be also appropriate. Since labeled datasets of this specific historical manuscript do not exist, an expert took 56 lines belonging to 4 pages and annotated the containing 1583 symbols to be used for fine tuning. Of course, when using real labelled data, the results are better (a SER of 0.05) than without any fine tuning (a SER of 0.40). When we compare the quality of our produced-pseudo labels against the manually created ones, we observe that, by using pseudo-labeling, we reach a competitive result of 0.09 SER. This demonstrates the suitability of our method, because the performance is close to the one obtained with manual labels, while significantly reducing the annotation effort.

% \begin{table}[h]
% \centering
% \caption{Obtained Results on the Borg dataset. FT: Fine Tuning. Om: Omniglot. SD: Synthetic Data. RLD: Real Labels Data. PLD: Pseudo Labeled Data.}
% \label{tab:result_runic}   
% \noalign{\smallskip}

% \begin{tabular}{cccc}
% \hline\noalign{\smallskip}
% Method&User Effort & Training $\rightarrow$ FT & SER \\
% \noalign{\smallskip}\hline\noalign{\smallskip}
% \cite{souibgui2020}&Man. Labeling & Om $\rightarrow$ RLD & 0.05 \\
% \cite{souibgui2020}&5 shots & Om $\rightarrow$ NONE & 0.40 \\
% % provide 5 shots & Omniglot & BS & 0.28 \\
% Ours&5 shots & Om $\rightarrow$ SD $\rightarrow$ PLD & 0.09 \\
% Ours&5 shots & Om $\rightarrow$ SD + PLD & 0.09 \\
%  \noalign{\smallskip}\hline\noalign{\smallskip}
% \end{tabular}
% \end{table}

All in all, we can conclude that our proposed pseudo-labeling method achieves good results when recognizing low resource handwritten texts, with an important decrease in the user effort for data annotation. The analysis of the human effort is detailed next.

\subsection{Annotation Time Consumption}

Manually annotating data is a time consuming task and should be taken into account when using HTR models. Thus, in this section, we measure the time needed to label the three datasets to illustrate the manual labeling effort. As shown in Table~\ref{tab:time}, the more lines and the bigger the alphabet size, the more time is required to label the symbols with their bounding boxes. For reference, we measured the required time for providing the shots for our method and compared it with the manual annotation time. We found that locating and cropping 5 examples of each symbol in the alphabet takes approximately 40 seconds. Thus the user needed to spend 16 minutes for Borg, 17 min for Runicus and 52 min for Copiale for providing the shots for our approach. 
So, we can conclude that automatically providing pseudo-labels significantly minimizes the manual effort with a minimal loss in recognition performance compared to the manual annotation.
\begin{table}[h]
\centering
\caption{Required time (in minutes) for manually annotating the training lines.}
\label{tab:time}
\begin{tabular}{lcccc}
\hline\noalign{\smallskip}
Dataset & \# Lines&\# Symbols&\# Classes& Time \\
\noalign{\smallskip}\hline\noalign{\smallskip}
Borg & 117 & 1913&24 & $\approx$ 245 \\
Copiale & 176 &7197 &78 & $\approx$ 450  \\
Runicus & 56 & 1583&25 & $\approx$ 206 \\

 \noalign{\smallskip}\hline\noalign{\smallskip}
\end{tabular}
\end{table}

\subsection{Pseudo-labeling Performance Analysis}

Our proposed method progressively labels the dataset: we start by labelling easy symbols and progressively label the complicated ones. As a consequence, the accuracy of correctly labeling bounding boxes decreases as we select new pseudo labels at each iteration. We evaluated the quality of our pseudo-labeling approach on the three tested datasets by comparing the predicted bounding boxes and their corresponding pseudo-labels to the manually annotated ones. A predicted bounding box is considered is defined as a correct detection if it has a minimum overlap (Intersection over Union: IoU) of 0.7 with the ground-truth box.
We found that, more the dataset is difficult (in terms of segmentation, alphabet size and similarity between symbols), more the performance of our pseudo-labeling approach decreases and more iterations in the labeling process are needed. %, and vice versa.  
For example, the Borg labeling accuracy reaches 74 \% after obtaining all the labels. In Copiale, where symbols are easy to segment, the result was 85 \%. In Codex Runicus, we obtain the higher pseudo-labeling accuracy (a 94 \%) because the symbol segmentation is easier than Borg and the number of classes is lower than in Copiale. 

It is worth to mention that, during our experiments, we found that it is better to continue the labeling process despite a decreasing in the labeling performance. The reason is that, although we might add some wrong labels, in general, the incorporation of hard examples benefits the training and even a bounding box with a wrong label is still helping in the segmentation part. Moreover, the experiments show that there is a low difference between the manually annotated labels and our automatic produced ones, which encourages us for further improvements in our labeling process. 

\subsection{{Threshold Selection Study}}
{In our experiments we set a threshold of 0.4 before adding a character into the labeled set. This threshold is chosen after testing other values and finding that 0.4 is the optimal one. We show the results of the conducted experience in Table~\ref{tab:labelingthreshold}, where we tested different thresholds to select the pseudo-labels. The experiments were done on the Borg dataset.  }
 
\begin{table}[h]
\centering
\caption{{The symbol error rate when using different selecting thresholds while pseudo-labeling the data.}}
\label{tab:labelingthreshold}
\begin{tabular}{cc}
\hline\noalign{\smallskip}
Threshold & SER \\
\noalign{\smallskip}\hline\noalign{\smallskip}
0.8 & 0.27 \\
0.6 & 0.25 \\
0.4 & 0.24 \\
0.2 & 0.25 \\
 \noalign{\smallskip}\hline\noalign{\smallskip}
\end{tabular}
\end{table}

\subsection{{Is semi-Supervised learning worth?}}

{In this paper, we address handwriting recognition in low resource scenarios. Means, when there is a few labeled or unlabeled data to train on. So far, we opt to use an unsupervised approach that starts from a few shots of the desired alphabet. However, the choice of labeling some real lines as a start and pseudo-labeling the rest is also possible. We tested this strategy as presented in Table~\ref{tab:semi}. The obtained results show that starting with 50 \% of labeled lines (i.e 58 lines) lead to a good result, which is even better than the normal supervised training with 117 lines, this is due to the curriculum learning that improves the model convergence. However when reducing the starting lines to 30 \% or 20 \% the performance decreases and becomes similar to starting from only a few shots. It is to note also that starting with more manually labeled lines means obviously reducing the size of the unlabeled lines to be pseudo labeled, which also decreases the training time. Overall, we can conclude that starting from a few shots is a better solution in terms of avoiding the costly annotation effort, since the SER is slightly affected (we obtain 0.24 as SER using our unsupervised pseudo-labeling).}

\begin{table}[h]
\centering
\caption{{The semi-supervised approach performance.}}
\label{tab:semi}
\begin{tabular}{ccc}
\hline\noalign{\smallskip}
Labeled lines & SER & Training Time (hours)\\
\noalign{\smallskip}\hline\noalign{\smallskip}
20 \% & 0.25 &$\approx 23$\\
30 \% & 0.24 &$\approx 18$\\
50 \% & 0.20 &$\approx 17$\\
 \noalign{\smallskip}\hline\noalign{\smallskip}
\end{tabular}
\end{table}

\section{Conclusion}\label{sec:conclusion}

In this paper, we presented a novel pseudo-labeling few-shot transcription method for low-resource scenarios (manuscripts with rare alphabets and very few labelled data). We show that we can significantly reduce the human labor of annotating  handwritten datasets, while maintaining the performance. The performed experiments on the enciphered and historical manuscripts confirmed the usefulness, with a significant reduce in user effort and a minimal loss in recognition performance. 

Our pseudo-labeling few-shot model is a significant extension of our previous work \cite{souibgui2020}. In fact, its simplicity makes it even applicable on top of other methods, like \cite{zhang2020}. Also, for common scripts but with few labeled data, pseudo-labels can be predicted to train usual HTRs, which may lead to better results than the few-shot ones. 

As future work, we aim to enhance the quality of the provided labels to keep reducing the need of manual intervention. Also, we plan to extend our approach to work at paragraph or page level. It can be extended also to cover more low resource and other scripts.

\section*{Acknowledgement}
\noindent This work has been supported by the Swedish Research Council, grant 2018-06074, DECRYPT--Decryption of Historical Manuscripts, the Spanish project RTI2018-095645-B-C21 and the CERCA Program / Generalitat de Catalunya.
On behalf of Project DECRYPT we thank for the usage of MTA Cloud (https://cloud.mta.hu/) that significantly helped us achieve the results published in this paper.

\bibliography{main}
\end{document}